\documentclass[a4paper]{article}

\usepackage{INTERSPEECH2020}
\usepackage[utf8]{vietnam}
\usepackage[english]{babel}

\title{Improving Vietnamese Named Entity Recognition from Speech \\ Using Word Capitalization and Punctuation Recovery Models}
\name{Thai Binh Nguyen$^1$, Quang Minh Nguyen$^1$, Thi Thu Hien Nguyen$^{1,2}$,\\ Quoc Truong Do$^1$, Chi Mai Luong$^{1,3}$}

\address{
  $^1$Vietnam Artificial Intelligence System, Vietnam\\
  $^2$Thai Nguyen University of Education, Vietnam\\
$^3$University of Science and Technology of Hanoi, Vietnam
}
\email{\{binhnguyen|minhnguyen|truongdo\}@vais.vn, hienntt.math@tnue.edu.vn, lcmai@ioit.ac.vn}

\begin{document}

\baselineskip 10pt

\maketitle
\begin{abstract}
%  Study on Automatic Speech Recognition (ASR) and NamedEntity Recognition (NER) have shown outstanding results thatreach human parity. Notwithstanding that, researching the com-bination of two methods to handle problem NER in speech stillhas much more limited because of the difference between inputand output of two problems. In this research, we bring togetherprevious studies in two fields ASR and NER and come up withsome techniques that arm smoothly between these modules tohandle problem NER in speech. Ours main contributions are:(1) Prepare the first Vietnamese Speech NER dataset, (2) Ap-plying the punctuation and capitalization recovering model toimprove accurate of entity extraction on speech, (3) Proposinga technique to handle foreign language since that appears manytimes and have a big impact in the accurate of NER module,(4) Pretrained the first public large-scale monolingual languagemodels for Vietnamese that achieve the state-of-the-art in Viet-namese NER task. Experiment on the proposed dataset, our pro-posed approach achieves the improvement of more than 15%absolute change in the F1 score
 Studies on the Named Entity Recognition (NER) task have shown outstanding results that reach
 human parity on input texts with correct text formattings, such as with proper punctuation and capitalization. 
 However, such conditions are not available in applications where the input is speech, because the text is generated from a speech recognition system (ASR),
 and that the system does not consider the text formatting.
 In this paper, we (1) presented the first Vietnamese speech dataset for NER task, and (2) the first pre-trained public large-scale monolingual language  model for Vietnamese that achieved the new state-of-the-art for the Vietnamese NER task by 1.3\% absolute F1 score comparing to the latest study.
 And finally, (3) we proposed a new pipeline for NER task from speech that overcomes
 the text formatting problem by introducing a text capitalization and punctuation recovery model (CaPu) into the pipeline.
 The model takes input text from an ASR system and performs two tasks at the same time, producing proper text formatting that helps to improve NER performance.
 Experimental results indicated that the CaPu model helps to improve by nearly 4\% of F1-score.
 
 %Ours main contributions are: (1) Prepare the first Vietnamese Speech NER dataset, (2) Applying the punctuation and capitalization  recovering model to improve accurate of entity extraction on speech, (3) Proposing a technique to handle foreign language since that appears many times and have a big impact in the accurate of NER module, (4) Pretrained the first public large-scale monolingual language models for Vietnamese that achieve the state-of-the-art in Vietnamese NER task. Experiment on the proposed dataset, our proposed approach achieves the improvement of more than 15\% absolute change in the F1 score.

\end{abstract}

\section{Introduction}
Named Entity Recognition firstly has introduced in MUC-6 \cite{grishman1996message} in 1995, since then NER is a very active researched topic. Almost studies were done using written text, meanwhile, spoken text (that output from ASR) is much harder. One of the most challenges of doing NER from speech is dealing with the mismatch between ASR output and NER input. The mismatch is showing in many forms like do not have punctuation, not case sensitive. Sometimes, ASR error makes the meaning of the speech transcription output is changed that affecting the NER module harder processing accuracy. In \cite{inproceedingssurdeanu,articlekubala} evinced a significant improvement of NER result when the input is case-sensitive. Of course not any language need this kind of information like in \cite{inproceedingssudoh} doing NER in Japanese speech data, \cite{zhai-etal-2004-using} in Chinese language. 

Many types of architectural models were study in this problem can be referred as hidden markov models was used in study \cite{articlekubala,miller-etal-2000-named,inproceedingskim}, CRF in \cite{inproceedingsparada}, SVM in \cite{inproceedingssurdeanu,inproceedingssudoh} and maximum entropy model in \cite{zhai-etal-2004-using}. For now, deep learning model have proved as strong method for extracted entity, the study in \cite{Yadav2018ASO} showing this overwhelming. To increase the accuracy of extracting entity from speech, some studies not only using text output from the ASR module but also consider the speech features. The author in \cite{inproceedingsbechet} propose a 2-step approach for extracting named entity values. Thereby, first step the 1-best hypothesis produced by the ASR system will go through NER module to tag entity. Then the second step will extract the entity values from the word lattice on the areas selected by the named entity module. Another method for this strategy is in \cite{articlegoel} that adding loss of the NER model to loss of the ASR model so they can optimize the ASR model directly for the NER task. The research in this way needs to gather more data that have to include both text transcription and tagged entity in there. This makes collecting enough data to train models effectively more difficult, especially in rare languages. 

% Handling a foreign language is one big challenge in this problem. Almost ASR systems were trained in mono language, then when it have to handle a word in another language, they will try to output of some word of their language that have the same reading convention with that word. This causes a serious error and makes the meaning of the sentence nearly changed, consequence the NER module confused. The research in \cite{inproceedingsparada} proposed a method using a custom ASR model (propose in \cite{inproceedingscarolina}) to output an OOV word every time it catches a rare word. Then the NER module can learn to handle the OOV word just by using the context of the words but not care about the meaning of the OOV word. This method encountered a serious problem is "what is the real value of OOV?", because in the end, what we need is finger out which entity in speech.

The ASR output (hypothesis output) sometimes different from the reference text (due to ASR error) that makes hard to evaluate directly the accuracy of NER output. To fix the mismatch between these, many approaches was proposed. In \cite{articlekubala}, the authors make a module to align between hypothesis text and reference text before calculate the F1-score of NER result. The research in \cite{galibert2011structured} define other structured NER dataset (including hierarchical and compositional) for effective benchmark. But \cite{galibert2011structured} require re-label the NER dataset. In \cite{burger1998named} study, they upgraded of MUC score with phonetic alignment and content comparison. This approach, semantically, make the score closer to the actual accuracy.

In summary, we gain state-of-the-art in the NER VLSP 2018 dataset by using our own language model trained using the large Vietnamese corpus dataset. We also proposed a pipeline for extracting entities in the speech dataset, building a new dataset to evaluate this proposed pipeline and proves the effectiveness of the proposition. We release our pretraining code and model implemented in PyTorch.

% For the rest of this paper, section \ref{related-work} will survey related studies around subproblems in our system. Section \ref{proposed-method} will detail our system components and model architecture that we proposed. Data preparation and experiment result will be shown in section \ref{experiment}. Final section \ref{conclusion} will summarize our work and future development directions for this problem.

\section{Related works}\label{related-work}
\subsection{Automatic Speech Recognition}

The goal of Automatic Speech Recognition (ASR) is to address the transcription correlates with audio signals. Normally, an ASR system contains two main components, which are acoustic model and language model. 
%The most probable output sentence \(W\) given acoustic feature vector \(O\) is computed as below:
%\[\hat{W} = argmax\underbrace{P(O|W)}_\text{observation likelihood}\underbrace{P(W)}_\text{prior} \]
%The prior, \(P(W)\), is computed by the \textit{language model}; meanwhile the likelihood \(P(O|W)\) is computed by \textit{acoustic model}.

The main idea of language model (LM) comes from the need of predicting probability of next word. The n-gram model is frequently integrated into the ASR system to assign the conditional sequence of words. The LM in this paper is trained on large amount of text with various domain but mostly news. After training stage, the acoustic model finally composes with language model using a mapping dictionary of words and theirs pronunciation called lexicon.

\subsection{Capitalization and Punctuation Recovery}
Normally, the ASR output is lower case without any sentence mark, Capitalization and Punctuation Recovery (CaPu) is a task to regain that information. This is immensely important for other downstream tasks that use the output of ASR for example Machine Translation, NER. These tasks require correct text formattings, such as with proper punctuation and capitalization as one of the indispensable things.

Many studies on CaPu task for ASR processed in recently year. For example, \cite{cho2012segmentation} uses a phrase-based machine translation model to insert punctuation. The research in \cite{zelasko2018punctuation, tilk2015lstm} combine information from speech signal with text transcript to improve the performance. Conditional Random Field model was applied in \cite{lu2010better, ueffing2013improved} to insert punctuation that considered as a part of speech tagging problem. Some of the newest works are \cite{capubinh, capuhien} that using the Transformer model applies with the chunk merging technique to do the CaPu task. This technique showing very effective because it processes with the output of ASR without boundary. In our work, we use this approach to handle CaPu task.

\subsection{Named Entity Recognition}
Aggarwal and Zhai state the definition of an entity's identification problem \cite{aggarwal2012mining}: the Name Entity Recognition function, also referred to as NER, is to recognise and classify named entities from free text into a collection of predefined categories, such as person, organisation, and location.
 
Highly effective entity recognition studies are also focused on statistics machine learning. One of the first machine learning methods applied to NER is the Hidden Markov model (HMM). Nymble is the first HMM-based NER system developed by Bikel et al \cite{bikel1998nymble}. Later built machine learning methods for NER have shifted from the HMM model to the discreet models to overcome the HMM disadvantages. A common discrete model used in entity recognition is the Maximum Entropy model (MaxEnt). In \cite{chieu2002named}. The results of the system were evaluated to be not inferior to those of other contemporaries but gained the advantage of model building costs. In \cite{mccallum2000maximum}, the authors proposed the maximum Markov Entropy model for the problem of entity identification. The typical studies for NER using this model \cite{bender2003maximum, curran2003language} have also achieved positive results. Conditional Random Fields (CRF) is another typical discrete model for entity identification problem \cite{mccallum2003early}. In \cite{sarawagi2005semi}, the authors proposed a semi-Markov Conditional Random Fields model and applied to NER, the results were assessed to be better than conventional CRFs.

Recently, studies have focused on improving the efficiency of entity identification by adding information generated from large amounts of unlabeled data. This approach is the application of semi-supervised machine learning and deep learning machine learning techniques \cite{collobert2011natural}. Since then, the deep learning method has been extensively studied for the NER problem. Many new models have been applied, improved, combined and achieved important results in many different languages \cite{ji2019deep, trandafili2020named}.

% \subsection{Named Entity Recognition in Speech}

\section{Proposed method}\label{proposed-method}
\subsection{System overview}\label{sys-overview}
\begin{figure}[t]
  \centering
  \includegraphics[width=0.75\linewidth]{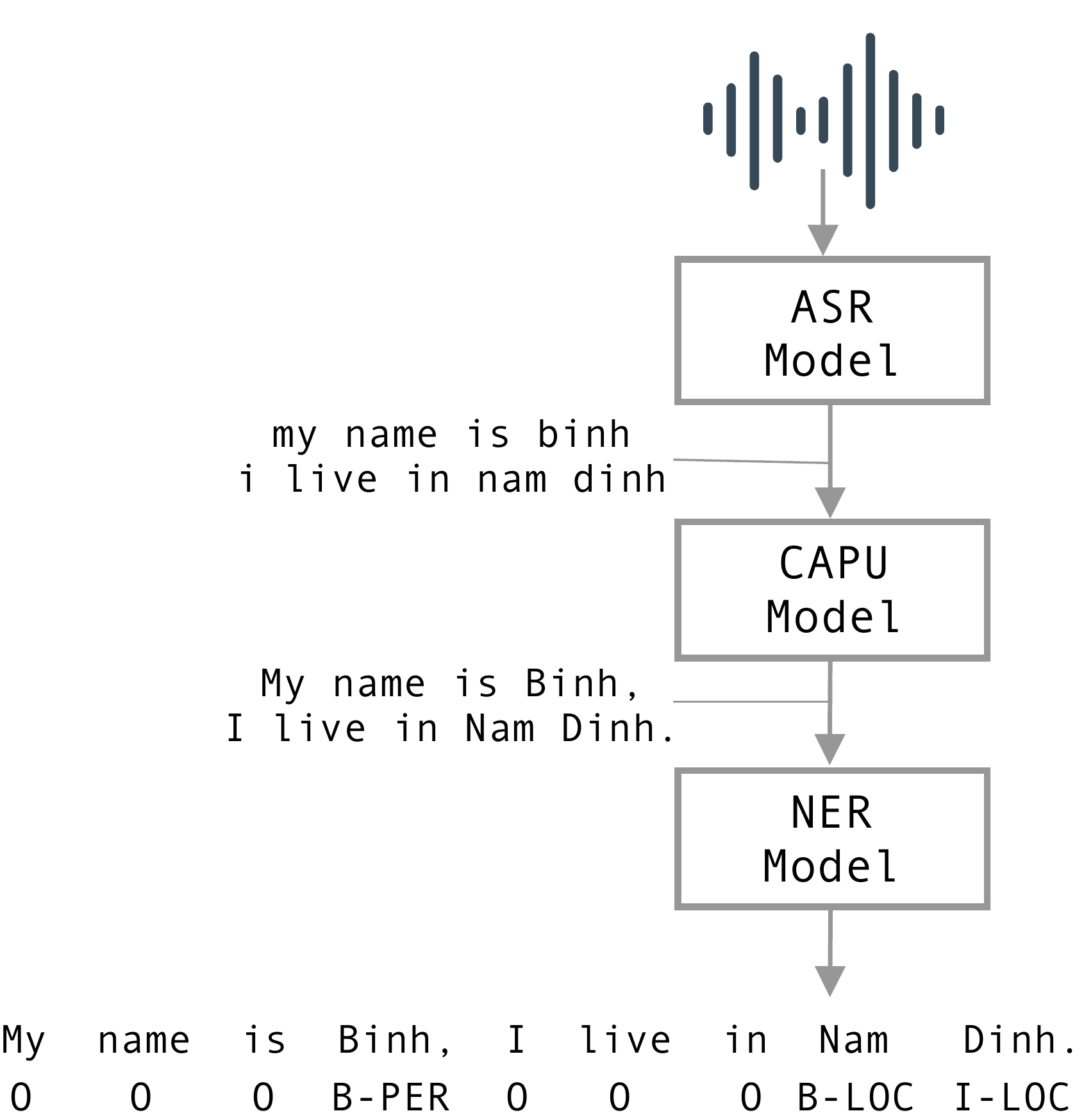}
  \caption{NER in speech system overview}
  \label{fig:sys-overview}
\end{figure}

Figure \ref{fig:sys-overview} show an overview of the system extract NER in Speech. This system has 3 distinct submodules respectively are ASR, CaPu, and NER. ASR model will transcript speech signal to text, this text in raw format that doesn't have punctuation and all lowercase. This text will be reformatted using the CaPu model and then information of entities is taken out from that text by using the NER model. 

% \subsection{Handling foreign words}
% Despite that an ordinary ASR system usually trained on a specify primary language, there are some minor amount of foreign words still exist. In this work, the primary language is Vietnamese. However, it is very popular that there are foreign person names, organization or location appear in the speaking content. Even though, acoustic model of ASR can resolve how those words pronounced but foreign words are out of vocabulary in language model then the system cannot give the correct foreign words in final result.

% To avoid these exceptional cases, there are two parts of ASR system must be updated: \textit{lexicon} and \textit{language model}. The vocabulary of general language model covers a wide range of words. Some of most frequent foreign words are filtered out in that corpus. By using a manually prepared dictionary mapping Vietnamese pronunciation and those words, these rare foreign words can be treated as normal words in primary language. 

\subsection{Capitalization and Punctuation Model}\label{capumodel}

Upper case and sentence mark take an important role that provides the meaning of a sentence, to be one of the indispensable information need to feed in a NER model. Unfortunately, this information was ignored in ASR model. Figure \ref{fig:sys-overview} reveals our proposal to attach a CaPu model to recovery this kind of information. 

\begin{figure}[t]
  \centering
  \includegraphics[width=1.05\linewidth]{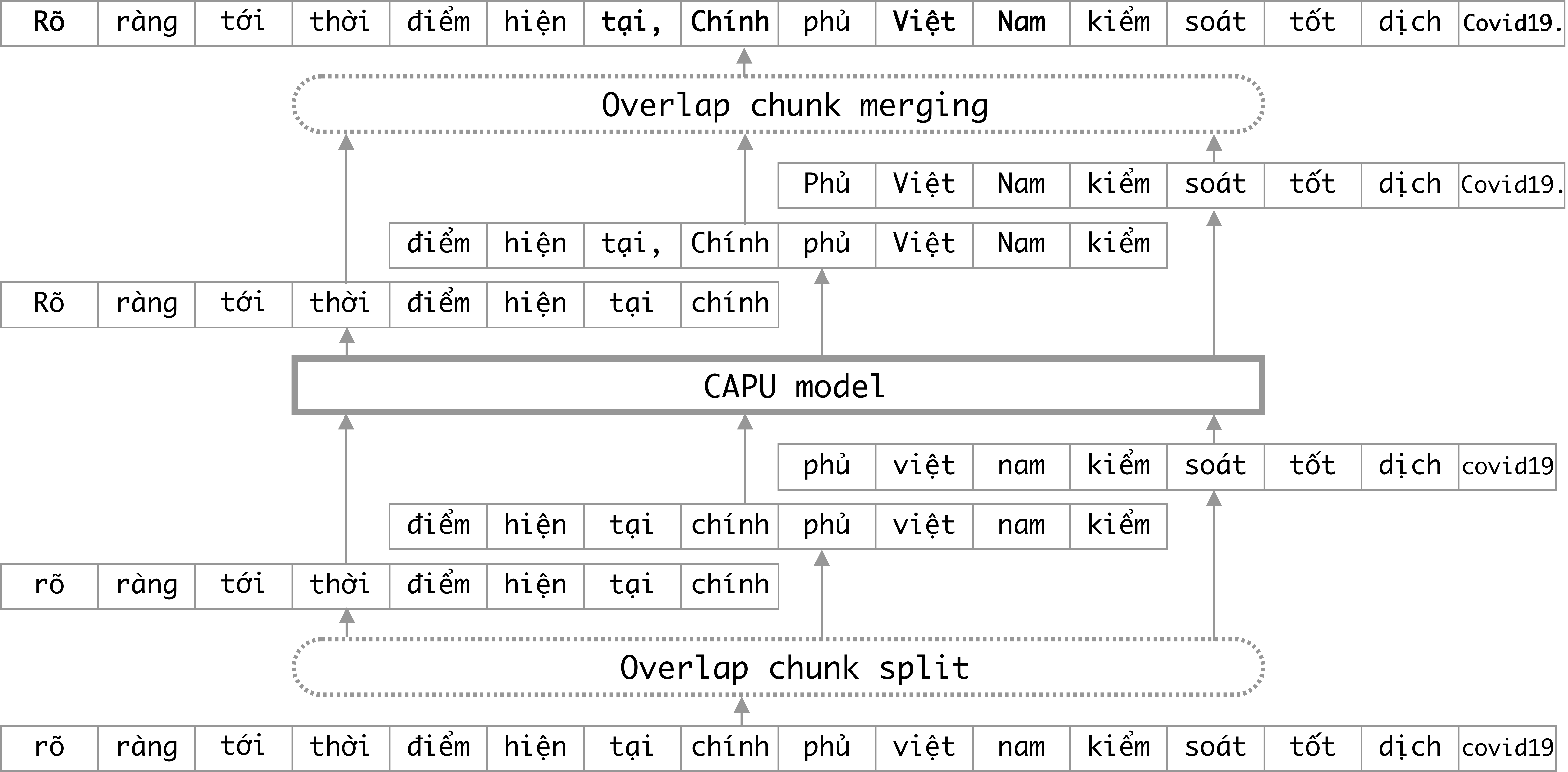}
  \caption{CAPU model apply chunk merging in Vietnamese \cite{capubinh}. Step 1: Split input sentence to overlap chunks. Step 2: Producing formatted text using CaPu model. Step 3: Concatenation the overlap chunks.}
  \label{fig:capu-example}
\end{figure}

The study in \cite{capubinh} shows very suitably for our task because it can handle endless string that output from the ASR model and having a good performance. In \cite{capubinh}, the authors proposed to handle the CaPu problem as a translation task that translate a lower case string without punctuation to a normal string with uppercase and sentence mark. The model was based using the Transformer sequence to sequence model and apply technique chunk merging. Example in figure \ref{fig:capu-example} shows the idea how the CaPu model working with chunk merging technique. The input text is split into three overlap chunks then transformed in parallel using CaPu model. The idea of merging overlap chunks is to replace the boundary words by the same words but in the middle of other sentences. Example \textbf{`tại chính'} words in the first sentence are replaced by \textbf{`tại, Chính'} words in the second sentence to produce the final output.

\subsection{Named Entity Recognition Model}
The main difference in our NER module in this system is that instead of training from scratch or using other pre-trained word embeddings, we train our own word embedding. Pre-trained language models, especially the Bidirectional Encoder Representations (BERT) recently become really popular and helped to produce significant improvement gains for various NLP tasks. 

\begin{figure}[t]
  \centering
  \includegraphics[width=0.8\linewidth]{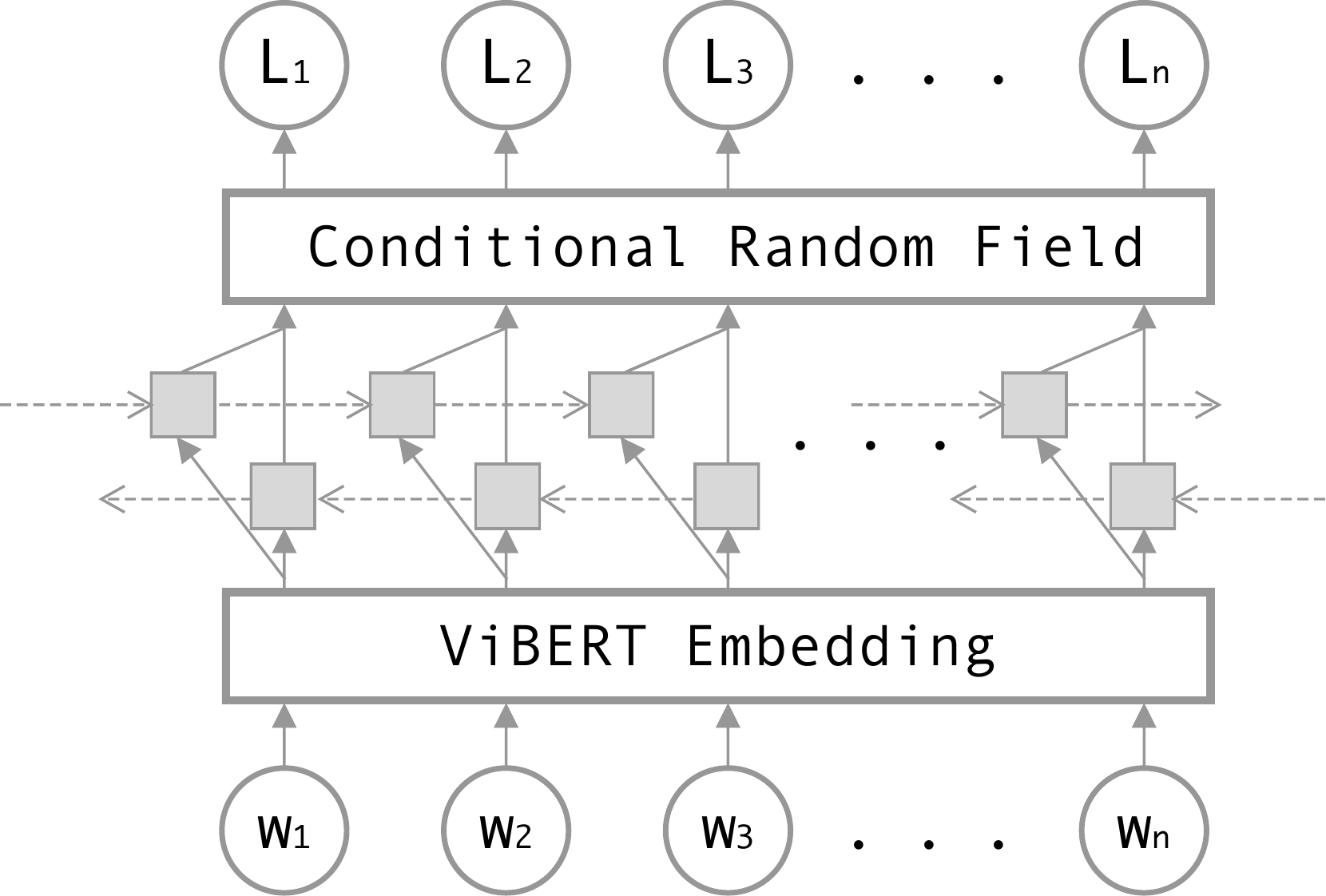}
  \caption{Named Entity Recognition Model}
  \label{fig:ner-model}
\end{figure}

In our work, we use RoBERTa architecture \cite{liu2019roberta} (an improved recipe for training BERT models) and train it on Vietnamese corpus to make pre-trained language model. Because of the limitation of computing resources, we reduce the number of hidden layers and attention head and embedding dimension from the RoBERTa base model architecture (detail in section \ref{model-setup}). The trained model we named it ViBERT. The pre-trained model we public in Github\footnote{https://github.com/nguyenvulebinh/vietnamese-roberta}. Figure \ref{fig:ner-model} depicts our NER model design. ViBERT was used to embed the input sentence. The bi-directional GRU models and CRF layer attached to the top of ViBERT to classify the entity-tag of each input word.

\section{Speech NER dataset} \label{speech-ner-dataset}

NER on speech is one of the problems with rare official data even in the most popular language like English. In Vietnamese, this is the first attempt to build a dataset for this task. Two approaches that we consider are creating speech data from the NER dataset or vice versa. Recording audio from reading the NER text reference is much easier than tagging NER on the ASR reference transcript. 

NER VLSP 2018 \cite{JCC13161} is the best dataset, for now, to benchmark an NER system in the Vietnamese language. Because of the limit of resources, we only consider recording audio reading the test set in the dataset that includes 4272 samples with total 242k words. The speech data was created by 4 people reading in different environments that produce a total of more than 26 hours of audio. 

\section{Experiments}\label{experiment}
In this section, we evaluate the effectiveness of the 
the proposed NER model with RoBERTa language models,
and how good the CaPu models can contribute to the NER accuracy on input speech data.

With regards to the ASR system, we utilized time-delayed
neural network models (TDNN) trained with the Kaldi toolkit \cite{povey2011kaldi}.
The model has 22 layers and is trained with 3000 hours of
speech data.

\subsection{Data preparation}\label{data-preparation}

NER dataset for the Vietnamese language was prepared in the Vietnamese Language and Speech Processing Conference (VLSP). It contains total 32.000 samples for the training set and 4272 samples for testing purpose. The entities need to be extracted are a person name (PER), name of an organizer (ORG) and location name (LOC). This dataset was detailed in \cite{JCC13161}. The original data in XML format and contains entities at nested levels. To make it facilitate comparing with the public the result in \cite{JCC13161}, data was converted to CoNLL NER format and just detecting the entity at the first level.

CaPu and ViBERT need a large corpus to train. We use 50GB of text with approximate 7.7 billion words that crawl from many domains on the internet including news, law, entertainment, wikipedia and so on. Because of the variety of typing encode in Vietnamese language on the internet, we use Visen \footnote{https://github.com/nguyenvulebinh/visen} library to unify encode method. 

In the CaPu problem, punctuations were handle include `\textbf{.} \textbf{,}'. We split the corpus into segments of random range from 4 to 60 words. The total data that use to train the CaPu model is more than 300 million samples. All data are prepared in the type of encoded as the description in \cite{capubinh}. To evaluate CaPu model, we use the reference text in VLSP 2018 test set.

% \cite{sennrich-etal-2016-neural}
ViBERT model was trained using corpus data processed by the byte-pair-encoding (BPE) algorithm. BPE was set up to output vocabulary size 50k. We don't use Vietnamese word segmentation because it makes the size of vocabulary bigger and it doesn't make better representation due to the transformer architecture already make good use context of the sentence with multi heads attention.

To training the ASR model, we prepare Vietnamese speech corpus consists of approximately 3000 hours of speech data including various domains and speaking styles. The data is augmented with noise and room impulse respond to increase the quantity and prevent over-fitting.

The dataset prepared in section \ref{speech-ner-dataset} was used to evaluate the performance of the whole system. This dataset includes speech data to evaluate the ASR system and entity-tag to evaluate the accuracy of extracted entity from the ASR output by using the NER model.

% The last dataset is English to Vietnamese Phonetic Alphabet

\subsection{Evaluation metric}

In the normal NER system, the input is text and the output is a label for each word in that text. However, in our system (extracting NER in Speech), the input is speech and we need extract every entities in that audio. As system description in section \ref{sys-overview}, NER module will extract entities from the output of the ASR module. The problem here is the ASR output can have some types of errors like insertions, deletions, substitutions that make the length of hypothesis output labels may be different from the ground truth labels, that make it impossible to calculate the \textit{F1} score like in the normal NER system. 

\begin{figure}[t]
  \centering
  \includegraphics[width=1.0\linewidth]{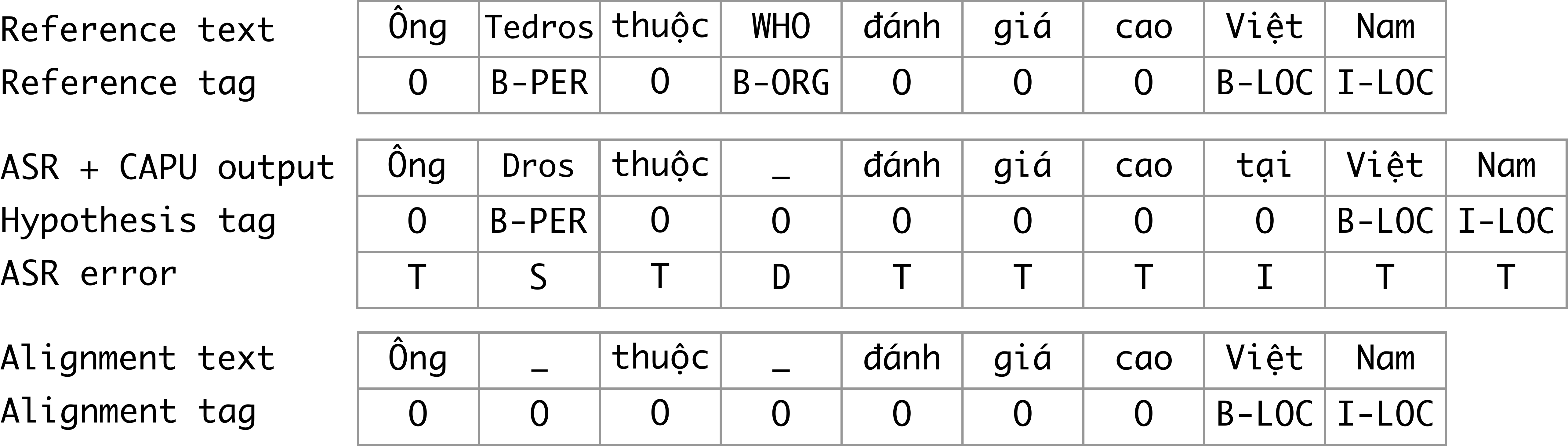}
  \caption{Evaluation example}
  \label{fig:system-example}
\end{figure}

To bypass this mismatch, the ASR output will be compared with the reference text in the NER dataset. If ASR output is right (\textbf{T}), the hypothesis entity tag was remained. If the error type is deletions (\textbf{D}) or substitutions (\textbf{S}), the hypothesis output of this word will become tag \textbf{O}. Else if error type is insertions (\textbf{I}), the label will be removed. That way will make the size of reference labels equal with the size of hypothesis labels. For example in figure \ref{fig:system-example} ASR output have some error. After alignment, the precision ($P$) is 100\% and recall ($R$) is 33.33\% because only 1 of 3 tags is correct (Việt Nam). $F1 = 2 * (P * R) / (P + R) = 50\%$ 

\subsection{Model setups}\label{model-setup}
%\subsubsection{ASR model setup}
%ABC
% \textbf{ASR model}

% \textbf{EnPhonVi model}

% In the CAPU model, word embedding size is 512, the number of transformer layer is 4 and each transformer component has 4 heads. After transform through transformer encoder block, all feature was input to a CRF layer and output 
% label for each input word.
%\subsubsection{ViBert model setup}
We reduce the size of the $RoBERTa_{base}$ model that implementation in fairseq \cite{ott2019fairseq} to produce ViBERT. This model contains 4 encoder layers compare with 4 layers in $RoBERTa_{base}$. The number of heads also reduce from 12 to 4 and the hidden dimension size is down from 768 to 512. Each training sample contains at most 512 subwords tokens. Like \cite{liu2019roberta}, we optimize the models using Adam.
ViBERT was trained using a batch size of 512 and a peak learning rate of 0.0003 with 3000 warming update steps. The total updating steps is 800k. We use 2 Nvidia 2080Ti GPUs (12GB each) for 5 weeks. 

% \subsubsection{NER model setup}
NER model setting use ViBERT for word representation and has 4 bidirectional GRU layers. The hidden size of GRU cell is 512. CRF was used in the output layer to produce 9 labels (\textbf{B-X}, \textbf{I-X}, \textbf{O} where \textbf{X} in set \{\textbf{ORG}, \textbf{PER}, \textbf{LOC}\}). This model also was optimized by using adam, the batch size is 64 and the training process converges after 30 epochs.

\subsection{Benchmark result}

% Table \ref{tab:asr-result} showing the testing result of the ASR model. The word error rate has a significant improvement in 4.1\% for absolute change that prove the process handles the foreign words work really well.

% \begin{table}[th]
%   \caption{Evaluation ASR model}
%   \label{tab:asr-result}
%   \centering
%   \begin{tabular}{ l  l }
%     \toprule
%     \multicolumn{1}{l}{\textbf{Model type}} &
%     \multicolumn{1}{l}{\textbf{Word error rate}} 
%     \\
%     \midrule
%     ASR normal system  & 10.67\% 
%     \\
%     ASR map English word system  & 6.57\% 
%     \\
%     \bottomrule
%   \end{tabular}
% \end{table}

Our NER model is the pre-trained word embedding (ViBERT) combine with GRU and CRF layer shows its effection making the SOTA result (90.18\%) when compared with the previous public result (table \ref{tab:ner-result}). This is the result evaluated directly using the reference text that is case sensitive and having full punctuation.

\begin{table}[th]
  \caption{Evaluation NER models on the NER VLSP 2018 dataset}
  \label{tab:ner-result}
  \centering
  \begin{tabular}{ l  l }
    \toprule
    \multicolumn{1}{l}{\textbf{Model type}} &
    \multicolumn{1}{l}{\textbf{F1 score}} 
    \\
    \midrule
    Vi Tokenizer + Bidirectional Inference \cite{JCC13161}  & 88.78\% 
    % \\
    % CRF [$\clubsuit$]  & 78.4\% 
    \\
    VNER \cite{nguyen2019attentive} & 77.52\%
    \\
    % BiLSTM + CRF [$\clubsuit$] & 83.25\% 
    % \\
    Multi layers LSTM \cite{JCC13161}  & 83.8\% 
    \\
    CRF/MEM+BS \cite{JCC13161} & 84.08\% 
    \\
    ViBERT + GRU + CRF (our result)  & \textbf{90.18\%} 
    \\
    \bottomrule
  \end{tabular}
\end{table}

The word error rate of our ASR system is 6.57\%. Table \ref{tab:speed} shows that if the ASR output put directly to the NER model, the entity identity results are critically reduced from 90.18\% to 63.89\%. The importance of the uppercase letter and punctuation also be observed in the experiment of running NER model on uncased reference text (without case sensitive and punctuation). In this case, F1 score drop vastly from 90.18\% to 75.35\%.

% \begin{table}[th]
%   \caption{Evaluation NER models on the different text input type}
%   \label{tab:speed}
%   \centering
%   \begin{tabular}{ l  l  l }
%     \toprule
%     \multicolumn{1}{l}{\textbf{Input type}} &
%     \multicolumn{1}{l}{\textbf{F1}} &
%     \multicolumn{1}{l}{\textbf{F1 partial}} 
%     \\
%     \midrule
%     ASR output & 51.98\% & 57.92\%
%     \\
%     ASR output map English word  & 63.89\% & 71.16\% 
%     \\
%     ASR output map English word + CAPU  & 67.13\%  & 78.34\% 
%     \\
%     Uncased reference text & 75.35\% & 83.77\%
%     \\
%     Uncased reference text + CAPU & 81.41\% & 86.41\%
%     \\
%     \bottomrule
%   \end{tabular}
% \end{table}

\begin{table}[th]
  \caption{Evaluation NER models on the different text input type}
  \label{tab:speed}
  \centering
  \begin{tabular}{ l  l }
    \toprule
    \multicolumn{1}{l}{\textbf{Input type}} &
    \multicolumn{1}{l}{\textbf{F1 score}}
    \\
    \midrule
    Reference text & 90.18\%
    \\
    ASR output  & 63.19\% 
    \\
    ASR output + CAPU  & 67.13\%
    \\
    Uncased reference text & 75.35\%
    \\
    Uncased reference text + CAPU & 81.41\%
    \\
    \bottomrule
  \end{tabular}
\end{table}

\begin{figure}[t]
  \centering
  \includegraphics[width=0.8\linewidth]{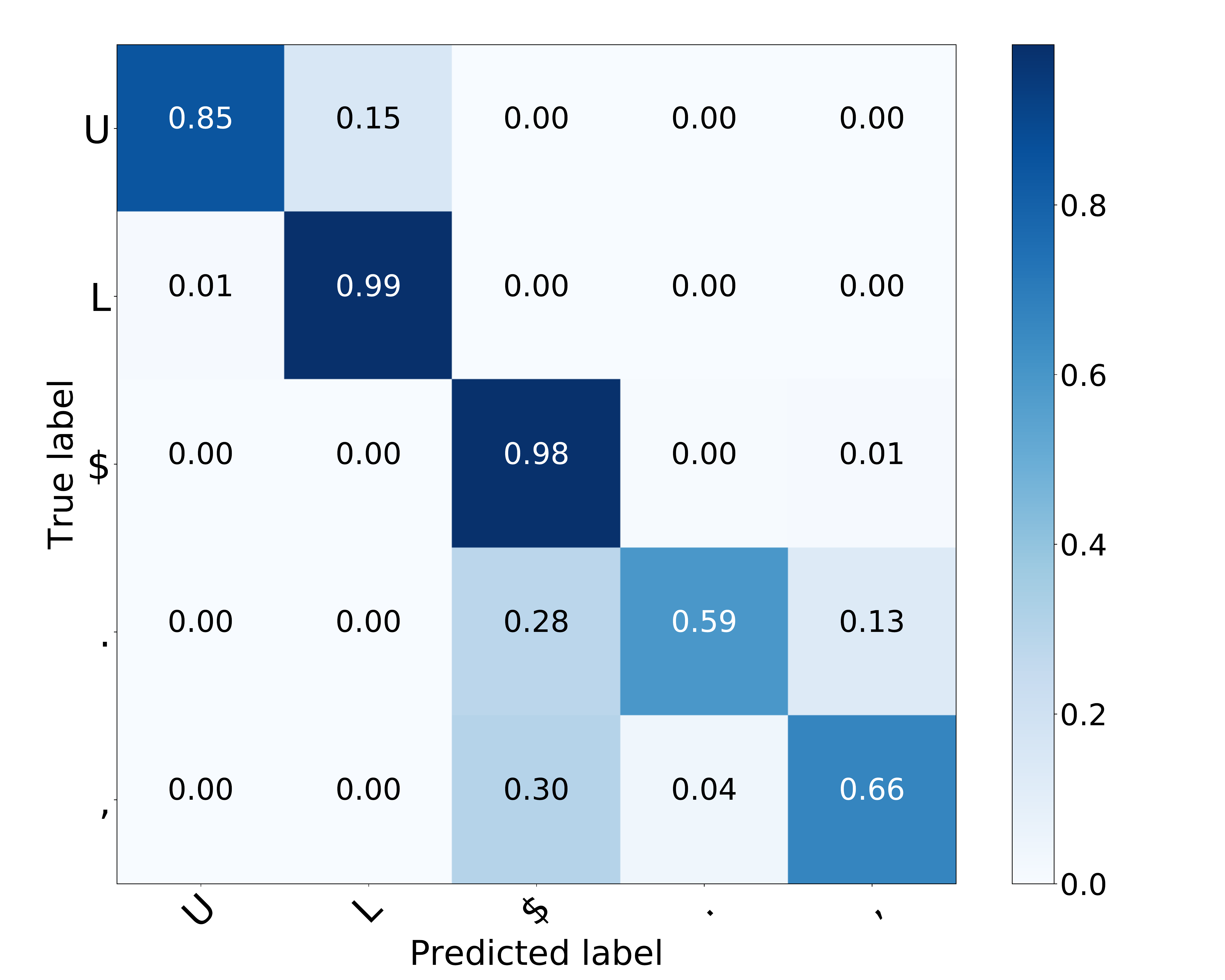}
  \caption{Evaluating the CAPU model on the uncased reference text (without case sensitive and punctuation)}
  \label{fig:capu-result}
\end{figure}

The main different between the ASR output and the reference text is the case sensitive and punctuation (ignore the error of the ASR model). In the figure \ref{fig:capu-result} demonstrate the result of the CaPu model. The accurate of recovery upper case letter is 85\%. Recuperation of punctuation is harder when the accurate is remain in nearly 60\% for the dot mark (\textbf{`.'}) and 66\% for comma mark \textbf{`,'}. The error of recovering punctuation happened when the CaPu model does not understand the meaning of the input sentence and put a blank mark (\textbf{\$}) after these words.

Table \ref{tab:speed} elucidate the effectiveness of the CaPu model in improving the accuracy of the NER model working on ASR output. F1 score of the NER model raises nearly 4\% from 63.19\% to 67.13\% when applying the CAPU model on the ASR output. The CaPu model also shows its value, improving more than 6\% F1 score of the NER model when apply to the uncased reference text.

\section{Conclusions}\label{conclusion}

In this study, we proposed to apply the capitalization and punctuation recovery model to improve named entity recognition from Vietnamese speech. The experimentation demonstrated the effectiveness of this combination. Through the research, we presented the first speech dataset that lay the groundwork for doing study extracted entity in speech for Vietnamese language. Besides that, we presented the effective impact of pre-trained language model for Vietnamese language apply for NER task that achieved new state-of-the-art on VLSP 2018 dataset. We public this pre-trained model to other researchers can further experiment. This is very useful for minority languages like Vietnamese. 

% Our proposed 

\bibliographystyle{IEEEtran}

\bibliography{mybib}

% Generated by IEEEtran.bst, version: 1.13 (2008/09/30)
\begin{thebibliography}{10}
\providecommand{\url}[1]{#1}
\csname url@samestyle\endcsname
\providecommand{\newblock}{\relax}
\providecommand{\bibinfo}[2]{#2}
\providecommand{\BIBentrySTDinterwordspacing}{\spaceskip=0pt\relax}
\providecommand{\BIBentryALTinterwordstretchfactor}{4}
\providecommand{\BIBentryALTinterwordspacing}{\spaceskip=\fontdimen2\font plus
\BIBentryALTinterwordstretchfactor\fontdimen3\font minus
  \fontdimen4\font\relax}
\providecommand{\BIBforeignlanguage}[2]{{%
\expandafter\ifx\csname l@#1\endcsname\relax
\typeout{** WARNING: IEEEtran.bst: No hyphenation pattern has been}%
\typeout{** loaded for the language `#1'. Using the pattern for}%
\typeout{** the default language instead.}%
\else
\language=\csname l@#1\endcsname
\fi
#2}}
\providecommand{\BIBdecl}{\relax}
\BIBdecl

\bibitem{grishman1996message}
R.~Grishman and B.~M. Sundheim, ``Message understanding conference-6: A brief
  history,'' in \emph{COLING 1996 Volume 1: The 16th International Conference
  on Computational Linguistics}, 1996.

\bibitem{inproceedingssurdeanu}
M.~Surdeanu, J.~Turmo, and E.~Comelles, ``Named entity recognition from
  spontaneous open-domain speech,'' in \emph{Ninth European Conference on
  Speech Communication and Technology}, 2005.

\bibitem{articlekubala}
F.~Kubala, R.~Schwartz, R.~Stone, and R.~Weischedel, ``Named entity extraction
  from speech,'' in \emph{Proceedings of DARPA Broadcast News Transcription and
  Understanding Workshop}.\hskip 1em plus 0.5em minus 0.4em\relax Citeseer,
  1998, pp. 287--292.

\bibitem{inproceedingssudoh}
K.~Sudoh, H.~Tsukada, and H.~Isozaki, ``Incorporating speech recognition
  confidence into discriminative named entity recognition of speech data,'' in
  \emph{Proceedings of the 21st International Conference on Computational
  Linguistics and the 44th annual meeting of the Association for Computational
  Linguistics}.\hskip 1em plus 0.5em minus 0.4em\relax Association for
  Computational Linguistics, 2006, pp. 617--624.

\bibitem{zhai-etal-2004-using}
\BIBentryALTinterwordspacing
L.~Zhai, P.~Fung, R.~Schwartz, M.~Carpuat, and D.~Wu, ``Using n-best lists for
  named entity recognition from {C}hinese speech,'' in \emph{Proceedings of
  {HLT}-{NAACL} 2004: Short Papers}.\hskip 1em plus 0.5em minus 0.4em\relax
  Boston, Massachusetts, USA: Association for Computational Linguistics, May 2
  - May 7 2004, pp. 37--40. [Online]. Available:
  \url{https://www.aclweb.org/anthology/N04-4010}
\BIBentrySTDinterwordspacing

\bibitem{miller-etal-2000-named}
\BIBentryALTinterwordspacing
D.~Miller, S.~Boisen, R.~Schwartz, R.~Stone, and R.~Weischedel, ``Named entity
  extraction from noisy input: Speech and {OCR},'' in \emph{Sixth Applied
  Natural Language Processing Conference}.\hskip 1em plus 0.5em minus
  0.4em\relax Seattle, Washington, USA: Association for Computational
  Linguistics, Apr. 2000, pp. 316--324. [Online]. Available:
  \url{https://www.aclweb.org/anthology/A00-1044}
\BIBentrySTDinterwordspacing

\bibitem{inproceedingskim}
J.-H. Kim and P.~C. Woodland, ``A rule-based named entity recognition system
  for speech input,'' in \emph{Sixth International Conference on Spoken
  Language Processing}, 2000.

\bibitem{inproceedingsparada}
C.~Parada, M.~Dredze, and F.~Jelinek, ``Oov sensitive named-entity recognition
  in speech,'' in \emph{Twelfth Annual Conference of the International Speech
  Communication Association}, 2011.

\bibitem{Yadav2018ASO}
V.~Yadav and S.~Bethard, ``A survey on recent advances in named entity
  recognition from deep learning models,'' in \emph{COLING}, 2018.

\bibitem{inproceedingsbechet}
F.~B{\'e}chet, A.~Gorin, J.~Wright, and D.~H. Tur, ``Named entity extraction
  from spontaneous speech in how may i help you?'' in \emph{Seventh
  International Conference on Spoken Language Processing}, 2002.

\bibitem{articlegoel}
V.~Goel and W.~Byrne, ``Task dependent loss functions in speech recognition:
  Application to named entity extraction,'' in \emph{ESCA Tutorial and Research
  Workshop (ETRW) on Accessing Information in Spoken Audio}, 1999.

\bibitem{galibert2011structured}
O.~Galibert, S.~Rosset, C.~Grouin, P.~Zweigenbaum, and L.~Quintard,
  ``Structured and extended named entity evaluation in automatic speech
  transcriptions,'' in \emph{Proceedings of 5th International Joint Conference
  on Natural Language Processing}, 2011, pp. 518--526.

\bibitem{burger1998named}
J.~D. Burger, D.~Palmer, and L.~Hirschman, ``Named entity scoring for speech
  input,'' in \emph{Proceedings of the 36th Annual Meeting of the Association
  for Computational Linguistics and 17th International Conference on
  Computational Linguistics-Volume 1}.\hskip 1em plus 0.5em minus 0.4em\relax
  Association for Computational Linguistics, 1998, pp. 201--205.

\bibitem{cho2012segmentation}
E.~Cho, J.~Niehues, and A.~Waibel, ``Segmentation and punctuation prediction in
  speech language translation using a monolingual translation system,'' in
  \emph{International Workshop on Spoken Language Translation (IWSLT) 2012},
  2012.

\bibitem{zelasko2018punctuation}
P.~{\.Z}elasko, P.~Szyma{\'n}ski, J.~Mizgajski, A.~Szymczak, Y.~Carmiel, and
  N.~Dehak, ``Punctuation prediction model for conversational speech,''
  \emph{arXiv preprint arXiv:1807.00543}, 2018.

\bibitem{tilk2015lstm}
O.~Tilk and T.~Alum{\"a}e, ``Lstm for punctuation restoration in speech
  transcripts,'' in \emph{Sixteenth annual conference of the international
  speech communication association}, 2015.

\bibitem{lu2010better}
W.~Lu and H.~T. Ng, ``Better punctuation prediction with dynamic conditional
  random fields,'' in \emph{Proceedings of the 2010 conference on empirical
  methods in natural language processing}, 2010, pp. 177--186.

\bibitem{ueffing2013improved}
N.~Ueffing, M.~Bisani, and P.~Vozila, ``Improved models for automatic
  punctuation prediction for spoken and written text.'' in \emph{Interspeech},
  2013, pp. 3097--3101.

\bibitem{capubinh}
B.~{Nguyen}, V.~B.~H. {Nguyen}, H.~{Nguyen}, P.~N. {Phuong}, T.~{Nguyen}, Q.~T.
  {Do}, and L.~C. {Mai}, ``Fast and accurate capitalization and punctuation for
  automatic speech recognition using transformer and chunk merging,'' in
  \emph{2019 22nd Conference of the Oriental COCOSDA International Committee
  for the Co-ordination and Standardisation of Speech Databases and Assessment
  Techniques (O-COCOSDA)}, 2019, pp. 1--5.

\bibitem{capuhien}
H.~N. {Thi Thu}, B.~N. {Thai}, H.~N. {Vu Bao}, T.~{Do Quoc}, M.~L. {Chi}, and
  H.~N. {Thi Minh}, ``Recovering capitalization for automatic speech
  recognition of vietnamese using transformer and chunk merging,'' in
  \emph{2019 11th International Conference on Knowledge and Systems Engineering
  (KSE)}, 2019, pp. 1--5.

\bibitem{aggarwal2012mining}
C.~C. Aggarwal and C.~Zhai, \emph{Mining text data}.\hskip 1em plus 0.5em minus
  0.4em\relax Springer Science \& Business Media, 2012.

\bibitem{bikel1998nymble}
D.~M. Bikel, S.~Miller, R.~Schwartz, and R.~Weischedel, ``Nymble: a
  high-performance learning name-finder,'' \emph{arXiv preprint
  cmp-lg/9803003}, 1998.

\bibitem{chieu2002named}
H.~L. Chieu and H.~T. Ng, ``Named entity recognition: a maximum entropy
  approach using global information,'' in \emph{Proceedings of the 19th
  international conference on Computational linguistics-Volume 1}.\hskip 1em
  plus 0.5em minus 0.4em\relax Association for Computational Linguistics, 2002,
  pp. 1--7.

\bibitem{mccallum2000maximum}
A.~McCallum, D.~Freitag, and F.~C. Pereira, ``Maximum entropy markov models for
  information extraction and segmentation.'' in \emph{Icml}, vol.~17, no. 2000,
  2000, pp. 591--598.

\bibitem{bender2003maximum}
O.~Bender, F.~J. Och, and H.~Ney, ``Maximum entropy models for named entity
  recognition,'' in \emph{Proceedings of the seventh conference on Natural
  language learning at HLT-NAACL 2003-Volume 4}.\hskip 1em plus 0.5em minus
  0.4em\relax Association for Computational Linguistics, 2003, pp. 148--151.

\bibitem{curran2003language}
J.~R. Curran and S.~Clark, ``Language independent ner using a maximum entropy
  tagger,'' in \emph{Proceedings of the seventh conference on Natural language
  learning at HLT-NAACL 2003}, 2003, pp. 164--167.

\bibitem{mccallum2003early}
A.~McCallum and W.~Li, ``Early results for named entity recognition with
  conditional random fields, feature induction and web-enhanced lexicons,'' in
  \emph{Proceedings of the seventh conference on Natural language learning at
  HLT-NAACL 2003-Volume 4}.\hskip 1em plus 0.5em minus 0.4em\relax Association
  for Computational Linguistics, 2003, pp. 188--191.

\bibitem{sarawagi2005semi}
S.~Sarawagi and W.~W. Cohen, ``Semi-markov conditional random fields for
  information extraction,'' in \emph{Advances in neural information processing
  systems}, 2005, pp. 1185--1192.

\bibitem{collobert2011natural}
R.~Collobert, J.~Weston, L.~Bottou, M.~Karlen, K.~Kavukcuoglu, and P.~Kuksa,
  ``Natural language processing (almost) from scratch,'' \emph{Journal of
  machine learning research}, vol.~12, no. Aug, pp. 2493--2537, 2011.

\bibitem{ji2019deep}
Y.~Ji, C.~Tong, J.~Liang, X.~Yang, Z.~Zhao, and X.~Wang, ``A deep learning
  method for named entity recognition in bidding document,'' in \emph{Journal
  of Physics: Conference Series}, vol. 1168, no.~3.\hskip 1em plus 0.5em minus
  0.4em\relax IOP Publishing, 2019, p. 032076.

\bibitem{trandafili2020named}
E.~Trandafili, E.~K. Me{\c{c}}e, and E.~Duka, ``A named entity recognition
  approach for albanian using deep learning,'' in \emph{Complex Pattern
  Mining}.\hskip 1em plus 0.5em minus 0.4em\relax Springer, 2020, pp. 85--101.

\bibitem{liu2019roberta}
Y.~Liu, M.~Ott, N.~Goyal, J.~Du, M.~Joshi, D.~Chen, O.~Levy, M.~Lewis,
  L.~Zettlemoyer, and V.~Stoyanov, ``Roberta: A robustly optimized bert
  pretraining approach,'' \emph{arXiv preprint arXiv:1907.11692}, 2019.

\bibitem{JCC13161}
\BIBentryALTinterwordspacing
H.~Nguyen, Q.~Ngo, L.~Vu, V.~Tran, and H.~Nguyen, ``Vlsp shared task: Named
  entity recognition,'' \emph{Journal of Computer Science and Cybernetics},
  vol.~34, no.~4, pp. 283--294, 2019. [Online]. Available:
  \url{http://vjs.ac.vn/index.php/jcc/article/view/13161}
\BIBentrySTDinterwordspacing

\bibitem{povey2011kaldi}
D.~Povey, A.~Ghoshal, G.~Boulianne, L.~Burget, O.~Glembek, N.~Goel,
  M.~Hannemann, P.~Motlicek, Y.~Qian, P.~Schwarz \emph{et~al.}, ``The kaldi
  speech recognition toolkit,'' in \emph{ASRU}, 2011.

\bibitem{ott2019fairseq}
M.~Ott, S.~Edunov, A.~Baevski, A.~Fan, S.~Gross, N.~Ng, D.~Grangier, and
  M.~Auli, ``fairseq: A fast, extensible toolkit for sequence modeling,'' in
  \emph{Proceedings of NAACL-HLT 2019: Demonstrations}, 2019.

\bibitem{nguyen2019attentive}
K.~A. Nguyen, N.~Dong, and C.-T. Nguyen, ``Attentive neural network for named
  entity recognition in vietnamese,'' in \emph{2019 IEEE-RIVF International
  Conference on Computing and Communication Technologies (RIVF)}.\hskip 1em
  plus 0.5em minus 0.4em\relax IEEE, 2019, pp. 1--6.

\end{thebibliography}
\end{document}